\documentclass[10pt,twocolumn,letterpaper]{article}

\usepackage{cvpr}
\usepackage{times}
\usepackage{epsfig}
\usepackage{graphicx}
\usepackage{amsmath}
\usepackage{amssymb}
\usepackage[misc]{ifsym}
\usepackage[bottom]{footmisc}
\usepackage[linesnumbered,ruled]{algorithm2e}

\usepackage[x11names,dvipsnames,table]{xcolor} %for use in color links
\usepackage{multirow}
\usepackage{array}
\usepackage[titletoc, title]{appendix}
\newcolumntype{L}[1]{>{\raggedright\let\newline\\\arraybackslash\hspace{0pt}}m{#1}}
\newcolumntype{C}[1]{>{\centering\let\newline\\\arraybackslash\hspace{0pt}}m{#1}}
\newcolumntype{R}[1]{>{\raggedleft\let\newline\\\arraybackslash\hspace{0pt}}m{#1}}

\usepackage[pagebackref=true,breaklinks=true,letterpaper=true,colorlinks,bookmarks=false]{hyperref}

 \cvprfinalcopy % *** Uncomment this line for the final submission

\def\cvprPaperID{****} % *** Enter the CVPR Paper ID here

\ifcvprfinal\fi
\begin{document}

%%%%%%%%% TITLE
\title{Snapshot Distillation: Teacher-Student Optimization in One Generation}

\author{Chenglin Yang\textsuperscript{1}\quad Lingxi Xie\textsuperscript{1}\quad Chi Su\textsuperscript{2}\quad Alan L. Yuille\textsuperscript{1}\quad\\
\textsuperscript{1}Johns Hopkins University\; \textsuperscript{2}Kingsoft\\
{\tt\small \{chenglin.yangw,198808xc,alan.l.yuille\}@gmail.com \quad suchi@kingsoft.com}
\and
%Lingxi Xie\\
%Johns Hopkins University\\
%{\tt\small 198808xc@gmail.com}
}

\maketitle
%\thispagestyle{empty}

%%%%%%%%% ABSTRACT
\begin{abstract}
Optimizing a deep neural network is a fundamental task in computer vision, yet direct training methods often suffer from over-fitting. Teacher-student optimization aims at providing complementary cues from a model trained previously, but these approaches are often considerably slow due to the pipeline of training a few generations in sequence, i.e., time complexity is increased by several times.

This paper presents snapshot distillation (SD), the first framework which enables teacher-student optimization in one generation. The idea of SD is very simple: instead of borrowing supervision signals from previous generations, we extract such information from earlier epochs in the same generation, meanwhile make sure that the difference between teacher and student is sufficiently large so as to prevent under-fitting. To achieve this goal, we implement SD in a cyclic learning rate policy, in which the last snapshot of each cycle is used as the teacher for all iterations in the next cycle, and the teacher signal is smoothed to provide richer information. In standard image classification benchmarks such as CIFAR100 and ILSVRC2012, SD achieves consistent accuracy gain without heavy computational overheads. We also verify that models pre-trained with SD transfers well to object detection and semantic segmentation in the PascalVOC dataset.
\end{abstract}

%%%%%%%%% BODY TEXT
\section{Introduction}
\label{Introduction}

A large portion of recent advances in computer vision have been built upon deep learning, in particular training very deep neural networks. With the depth increasing from tens \cite{krizhevsky2012imagenet}\cite{simonyan2015very}\cite{szegedy2015going} to hundreds~\cite{he2016deep}\cite{huang2017densely}, the issue of network optimization becomes a more and more important yet challenging problem, in which researchers proposed various approaches to deal with both under-fitting~\cite{nair2010rectified}, over-fitting~\cite{srivastava2014dropout} and numerical instability~\cite{ioffe2015batch}.

As an alternative approach to assist training, teacher-student (T-S) optimization was originally designed for training a smaller network to approximate the behavior of a larger one, {\em i.e.}, model compression~\cite{hinton2015distilling}, but later researcher found its effectiveness in providing complementary cues to training the same network~\cite{furlanello2018born}\cite{bagherinezhad2018label}. These approaches require a {teacher model} which is often obtained from a standalone training process. Then, an extra loss term which measures the similarity between the teacher and the student is added to the existing cross-entropy loss term. It was believed that such an optimization process benefits from so-called {\em secondary information}~\cite{yang2018knowledge}, {\em i.e.}, class-level similarity that allows the student not to fit the one-hot class distribution. Despite their success in improving recognition accuracy, these approaches often suffer much heavier computational overheads, because a sequence of models need to be optimized one by one. A training process with one teacher and $K$ students requires $K\times$ more training time compared to a single model.

\newcommand{\colwidthA}{0.8cm}
\begin{table}
\centering
{\setlength{\tabcolsep}{0.08cm}
\begin{tabular}{|l|C{\colwidthA}|C{\colwidthA}|C{\colwidthA}|}
\hline
{}                                                        & SA?        & IN?        & $1$G?      \\
\hline\hline
Knowledge Distillation (2015)~\cite{hinton2015distilling} &            &            &            \\
\hline
FitNet (2015)~\cite{romero2015fitnets}                    &            &            &            \\
\hline
Net2Net (2016)~\cite{chen2016net2net}                     &            & \checkmark &            \\
\hline
A Gift from KD (2017)~\cite{yim2017gift}                  &            &            &            \\
\hline
Label Refinery (2018)~\cite{bagherinezhad2018label}       & \checkmark & \checkmark &            \\
\hline
Born-Again Network (2018)~\cite{furlanello2018born}       & \checkmark &            &            \\
\hline
Tolerant Teacher (2018)~\cite{yang2018knowledge}          & \checkmark & \checkmark &            \\
\hline\hline
{\bf Snapshot Distillation} (this work)                   & \checkmark & \checkmark & \checkmark \\
\hline
\end{tabular}}
\caption{The attributes of different teacher-student optimization approaches, where SA indicates that teacher and student have the {\em same architecture}, IN indicates being evaluated on {\em ImageNet}, and $1$G indicates that the entire process is done within {\em one generation}. See Section~\ref{RelatedWork} for a detailed survey.}
\label{Tab:ApproachComparison}
\vspace{-0.1cm}
\end{table}

This paper presents an algorithm named {\bf snapshot distillation} (SD) to perform T-S optimization {\em in one generation} which, to the best of our knowledge, was not achieved in prior research. The differences between SD and previous methods are summarized in Table~\ref{Tab:ApproachComparison}. The key idea of SD is straightforward: taking extra supervision ({\em a.k.a.} the teacher signal) from the prior {\em iterations} (in the same generation) instead of the prior {\em generations}. Based on this framework, we investigate several factors that impact the performance of T-S optimization, and summarize three principles, namely, (i) the teacher model has been well optimized; (ii) the teacher and student models are sufficiently different from each other; and (iii) the teacher provides secondary information~\cite{yang2018knowledge} for the student to learn. Summarizing these requirements leads to our solution that using a cyclic learning rate policy, in which the last snapshot of each cycle (which arrives at a high accuracy and thus satisfies (i)), serves as the teacher for all iterations in the next cycle (these iterations are pulled away from the teacher after a learning rate boost, which satisfies (ii)). We also introduce a novel method to smooth the teacher signal in order to provide mild and more effective supervision (which satisfies (iii)).

Experiments are performed in two standard benchmarks for image classification, namely, CIFAR100~\cite{krizhevsky2009learning} and ILSVRC2012~\cite{russakovsky2015imagenet}. SD consistently outperforms the baseline (direct optimization) especially in deeper networks. In addition, SD requires merely $1/3$ extra training time beyond the baselines, which is much faster than the existing multi-generation approaches. We also fine-tune the models trained by SD for object detection and semantic segmentation in the PascalVOC dataset~\cite{everingham2010pascal} and observe accuracy gain, implying that the improvement brought by SD is transferrable.

The remainder of this paper is organized as follows. Section~\ref{RelatedWork} briefly reviews related work. Section~\ref{Approach} describes snapshot distillation and provides practical guides for T-S optimization in one generation. After experiments are shown in Section~\ref{Experiments}, we conclude this work in Section~\ref{Conclusions}.

\section{Related Work}
\label{RelatedWork}

Recently, the research field of computer vision has been largely boosted by the theory of deep learning~\cite{lecun2015deep}. With the availability of large-scale image datasets~\cite{deng2009imagenet} and powerful computational resources, researchers designed deep networks to replace traditional handcrafted features~\cite{perronnin2010improving} for visual understanding. The fundamental idea is to build a hierarchical network structure containing multiple {\em layers}, each of which contains a number of {\em neurons} having the same or similar mathematical functions, {\em e.g.}, convolution, pooling, normalization, {\em etc}. The strong ability of deep networks at fitting complicated feature-space distributions is widely verified in the previous literature. In a fundamental task known as image classification, deep convolutional neural networks~\cite{krizhevsky2012imagenet} have been dominating in the large-scale competitions~\cite{russakovsky2015imagenet}. To further improve classification accuracy, researchers designed deeper and deeper networks~\cite{simonyan2015very}\cite{szegedy2015going}\cite{he2016deep}\cite{huang2017densely}\cite{hu2018squeeze}, and also explored the possibility of discovering network architectures automatically~\cite{xie2017genetic}\cite{zoph2017neural}\cite{liu2018progressive}.

The rapid progress of deep neural networks has helped a lot of visual recognition tasks. Features extracted from pre-trained classification networks can be transferred to small datasets for image classification~\cite{donahue2014decaf}, retrieval~\cite{razavian2014cnn} or object detection~\cite{girshick2014rich}. To transfer knowledge to a wider range of tasks, researchers often adopt a technique named fine-tuning, which replaces the last few layers of a classification network with some specially-designed modules ({\em e.g.}, up-sampling for semantic segmentation~\cite{long2015fully}\cite{chen2016deeplab} and edge detection~\cite{xie2015holistically} or regional proposal extraction for object detection~\cite{girshick2015fast}\cite{ren2015faster}), so that the network can take advantage of the properties of the target problem while borrowing visual features from basic classification.

On the other hand, optimizing a deep neural network is a challenging problem. When the number of layers becomes very large ({\em e.g.}, more than $100$ layers), vanilla gradient descent approaches often encounter stability issues and/or over-fitting. To deal with them, researchers designed verious approaches such as ReLU activation~\cite{nair2010rectified}, Dropout~\cite{srivastava2014dropout} and batch normalization~\cite{ioffe2015batch}. However, as depth increases, the large number of parameters makes it easy for the neural networks to be over-confident~\cite{guo2017calibration}, especially in the scenarios of limited training data. An effective way is to introduce extra {\em priors} or {\em biases} to constrain the training process. A popular example is to assume that some visual categories are more similar than others~\cite{deng2010what}, so that a class-level similarity matrix is added to the loss function~\cite{verma2012learning}\cite{wu2017hierarchical}. However, this method still suffers the lack of modeling per-image class-level similarity ({\em e.g.}, a {\em cat} in one image may look like a {\em dog}, but in another image, it may be closer to a {\em rabbit}), which is observed in previous research~\cite{wang2014learning}\cite{akata2016label}\cite{zhang2018image}.

{\em Teacher-student optimization} is an effective way to formulate per-image class-level similarity. In this flowchart, a teacher student is first trained, and then used to guide the student network, in which process the output ({\em e.g.}, confidence scores) of the teacher network carries class-level similarity for each image. This idea was first proposed to distill knowledge from a larger teacher network and compress it to a smaller student network~\cite{hinton2015distilling}\cite{romero2015fitnets}, or initialize a deeper/wider network with pre-trained weights of a shallower/narrower network~\cite{chen2016net2net}\cite{simonyan2015very}. Later, it was extended in various aspects, including using an adjusted way of teacher supervision~\cite{szegedy2016rethinking}\cite{pereyra2017regularizing}, using multiple teachers towards a better guidance~\cite{tarvainen2017mean}, adding supervision to intermediate neural responses~\cite{yim2017gift}, and allowing two networks to provide supervision to each other~\cite{zhang2017deep}. Recently, researchers noted that this idea can be used to optimize deep networks in many {\em generations}~\cite{bagherinezhad2018label}\cite{furlanello2018born}, namely, a few networks with {\em the same architecture} are optimized one by one, in which the next one borrows supervision from the previous one. It was argued that the {\em softness} of the teacher signal plays an important role in educating a good student~\cite{yang2018knowledge}. Despite the success of these approaches in boosting recognition accuracy, they suffer from lower training efficiency, as in a $K$-generation process (one teacher and $K$ students) requires $K\times$ more training time. An inspiring cue comes from the effort of training a few models for ensemble within the same time~\cite{huang2018snapshot}, in which the number of iterations for training each model was largely reduced.

\section{Snapshot Distillation}
\label{Approach}

This section presents snapshot distillation (SD), the first approach that achieves teacher-student (T-S) optimization within one generation. We first briefly introduce a general flowchart of T-S optimization and build a notation system. Then, we analyze the main difficulties that limit its efficiency, based on which we formulate SD and discuss principles and techniques to improve its performance.

\subsection{Teacher-Student Optimization}
\label{Approach:Overview}

Let a deep neural network be ${\mathbb{M}}:{\mathbf{y}}={\mathbf{f}\!\left(\mathbf{x};\boldsymbol{\theta}\right)}$, where $\mathbf{x}$ denotes the input image, $\mathbf{y}$ denotes the output data ({\em e.g.}, a $G$-dimensional vector for classification with $G$ being the number of classes), and $\boldsymbol{\theta}$ denotes the learnable parameters. These parameters are often initialized as random noise, and then optimized using a training set with $N$ data samples, ${\mathcal{D}}={\left\{\left(\mathbf{x}_1,\mathbf{y}_1\right),\ldots,\left(\mathbf{x}_N,\mathbf{y}_N\right)\right\}}$.

Conventional optimization algorithm works by sampling mini-batches or subsets from the training set. Each of them, denoted as $\mathcal{B}$, is fed into the current model to estimate the difference between prediction and ground-truth labels:
\begin{equation}
\label{Eqn:StandardLoss}
{\mathcal{L}\!\left(\mathcal{B};\boldsymbol{\theta}\right)}={-\frac{1}{\left|\mathcal{B}\right|}{\sum_{\left(\mathbf{x}_n,\mathbf{y}_n\right)\in\mathcal{B}}}\mathbf{y}_n^\top\ln\mathbf{f}\!\left(\mathbf{x}_n;\boldsymbol{\theta}\right)}.
\end{equation}
This process searches over the parameter space to find the approximately optimal $\boldsymbol{\theta}$ that interprets or fits $\mathcal{D}$. However, the model trained in this way often over-fits the training set, {\em i.e.}, $\boldsymbol{\theta}$ cannot be transferred to the testing set to achieve good performance as in the training set. As observed in prior work~\cite{guo2017calibration}, this is partly because the supervision was provided in one-hot vectors, which forces the network to prefer the true class overwhelmingly to all other classes -- this is often not the optimal choice because rich information of class-level similarity is simply discarded~\cite{wu2017hierarchical}\cite{yang2018knowledge}.

To alleviate this issue, teacher-student (T-S) optimization was proposed, in which a pre-trained teacher network added an extra term to the loss function to measure the KL-divergence between teacher and student~\cite{furlanello2018born}:
\begin{eqnarray}
\label{Eqn:StudentLoss}
\nonumber
{\mathcal{L}^\mathrm{S}\!\left(\mathcal{B};\boldsymbol{\theta}^\mathrm{S}\right)}={-\frac{1}{\left|\mathcal{B}\right|}{\sum_{\left(\mathbf{x}_n,\mathbf{y}_n\right)\in\mathcal{B}}}\left\{\lambda^\mathrm{S}\cdot\mathbf{y}_n^\top\ln\mathbf{f}\!\left(\mathbf{x}_n;\boldsymbol{\theta}^\mathrm{S}\right)+\right.}\\
{\left.\lambda^\mathrm{T}\cdot\mathrm{KL}\!\left[\mathbf{f}\!\left(\mathbf{x}_n;\boldsymbol{\theta}^\mathrm{T}\right)\|\mathbf{f}\!\left(\mathbf{x}_n;\boldsymbol{\theta}^\mathrm{S}\right)\right]\right\}},
\end{eqnarray}
where $\boldsymbol{\theta}^\mathrm{S}$ and $\boldsymbol{\theta}^\mathrm{T}$ denote the parameters in teacher and student models, respectively. This is to say, the fitting goal of the student is no longer the ground-truth one-hot vector which is too strict, but leans towards the teacher signal (a softened vector most often with correct prediction). This formulation can be applied in the form of {\em multiple generations}. Let $K$ be the total number of generations~\cite{bagherinezhad2018label}\cite{furlanello2018born}\cite{yang2018knowledge}. These approaches started with a so-called patriarch model $\mathbb{M}^{\left(0\right)}$, and in the $k$-th generation, $\mathbb{M}^{\left(k-1\right)}$ was used to teach $\mathbb{M}^{\left(k\right)}$. \cite{yang2018knowledge} showed the necessity of setting {\em a tolerant teacher} so that the students can absorb richer information from class-level similarity and achieve higher accuracy.

Despite the ability of T-S optimization in improving recognition accuracy, it often suffers the weakness of being computationally expensive. Typically, a T-S process with one teacher and $K$ students costs $K\times$ more time, yet this process is often difficult to parallelize\footnote{To make fair comparison, researchers often train deep networks using a fixed number of GPUs. T-S optimization trains $K+1$ models serially, which is often difficult to accelerate even with a larger number of GPUs.}. This motivates us to propose an approach named {\bf snapshot distillation} (SD), which is able to finish T-S optimization {\em in one generation}.

\subsection{The Flowchart of Snapshot Distillation}
\label{Approach:Formulation}

The idea of SD is very simple. To finish T-S optimization in one generation, during the training process, we always extract the teacher signal from an earlier {\em iteration}, by which we refer to an intermediate status of the same model, rather than another model that was optimized individually.

Mathematically, let $\boldsymbol{\theta}_0$ be the randomly initialized parameters. The baseline training process contains a total of $L$ iterations, the $l$-th of which samples a mini-batch $\mathcal{B}_l$, computes the gradient of Eqn~\eqref{Eqn:StandardLoss}, and updates the parameters from $\boldsymbol{\theta}_{l-1}$ to $\boldsymbol{\theta}_l$. SD works by assigning a number ${c_l}<{l}$ for the $l$-th iteration, indicating a previous snapshot $\mathbf{f}\!\left(\mathbf{x};\boldsymbol{\theta}_{c_l}\right)$ as the teacher to update $\boldsymbol{\theta}_{l-1}$. Thus, Eqn~\eqref{Eqn:StudentLoss} becomes:
\begin{eqnarray}
\label{Eqn:SnapshotDistillation}
\nonumber
{\mathcal{L}\!\left(\mathcal{B}_l;\boldsymbol{\theta}_{l-1}\right)}={-\frac{1}{\left|\mathcal{B}_l\right|}{\sum_{\left(\mathbf{x}_n,\mathbf{y}_n\right)\in\mathcal{B}_l}}\left\{\lambda_l^\mathrm{S}\cdot\mathbf{y}_n^\top\ln\mathbf{f}\!\left(\mathbf{x}_n;\boldsymbol{\theta}_{l-1}\right)+\right.}\\
{\left.\lambda_l^\mathrm{T}\cdot\mathrm{KL}\!\left[\mathbf{f}\!\left(\mathbf{x}_n;\boldsymbol{\theta}_{c_l}\right)\|\mathbf{f}\!\left(\mathbf{x}_n;\boldsymbol{\theta}_{l-1}\right)\right]\right\}}.
\end{eqnarray}
Here $\lambda_l^\mathrm{S}$ and $\lambda_l^\mathrm{T}$ are weights for one-hot and teacher supervisions. When ${\lambda_l^\mathrm{T}}={0}$, the teacher signal is ignored at the current iteration, and thus Eqn~\eqref{Eqn:SnapshotDistillation} degenerates to Eqn~\eqref{Eqn:StandardLoss}. The pseudo code of SD is provided in Algorithm~\ref{Alg:SnapshotDistillation}. In what follows, we will discuss several principles required to improve the performance of SD.

\begin{algorithm}[t!]
\SetKwInOut{Input}{Input}
\SetKwInOut{Return}{Return}
\Input{
training set $\mathcal{D}$, number of iterations $L$, training configurations $\left\{\gamma_l,\lambda_l^\mathrm{T},\lambda_l^\mathrm{S},c_l\right\}_{l=1}^L$;
}
Initialize $\boldsymbol{\theta}_0$;\\
\For {${l}={1,2,\ldots,L}$} {
Sample a mini-batch $\mathcal{B}_l$ from $\mathcal{D}$;\\
Compute loss $\mathcal{L}\!\left(\mathcal{B}_l;\boldsymbol{\theta}_{l-1}\right)$ using Eqn~\eqref{Eqn:SnapshotDistillation};\\
${\boldsymbol{\theta}_l}\leftarrow{\boldsymbol{\theta}_{l-1}-\gamma_l\cdot\nabla_{\boldsymbol{\theta}_{l-1}}\mathcal{L}\!\left(\mathcal{B}_l;\boldsymbol{\theta}_{l-1}\right)}$
}
\Return{
${\mathbb{M}}:{\mathbf{y}}={\mathbf{f}\!\left(\mathbf{x};\boldsymbol{\theta}=\boldsymbol{\theta}_L\right)}$.
}
\caption{
Snapshot Distillation
}
\label{Alg:SnapshotDistillation}
\end{algorithm}

\subsection{Principles of Snapshot Distillation}
\label{Approach:Principles}

This subsection forms the core contribution of our work, which discusses the principles that should be satisfied to improve the performance of SD. In practice, this involves how to design the hyper-parameters $\left\{\gamma_l,\lambda_l,c_l\right\}_{l=1}^L$. We first describe three principles individually, and summarize them to give our solution in the final part.

\subsubsection{Principle \#1: The Quality of Teacher}
\label{Approach:Principles:TeacherQuality}

In prior work, the importance of having a high-quality teacher model has been well studied. At the origin of T-S optimization~\cite{hinton2015distilling}\cite{romero2015fitnets}\cite{yim2017gift}, a more powerful teacher model was used to guide a smaller and thus weaker student model, so that the teacher knowledge is distilled and compressed into the student. This phenomenon persists in a multi-generation T-S optimization in which teacher and student share the same network architecture~\cite{bagherinezhad2018label}.

Mathematically, the teacher model determines the second term on the right-hand side of Eqn~\eqref{Eqn:SnapshotDistillation}, {\em i.e.}, the KL-divergence between teacher and student. If the teacher is not well optimized and provides noisy supervision, the risk that two terms conflict with each other becomes high. As we shall see later, this principle is even more important in SD, as the number of iterations allowed for optimizing each student becomes smaller, and the efficiency (or the speed of convergence) impacts the final performance heavier. %This is also verified in large-scale experiments (see Section~\ref{Experiments:ILSVRC2012}), in which SD obtains higher accuracy gain when the baseline network is stronger.

\subsubsection{Principle \#2: Teacher-Student Difference}
\label{Approach:Principles:Difference}

In the context of T-S optimization in one generation, one more challenge emerges. In each iteration, the teacher $\boldsymbol{\theta}_{c_l}$ and student $\boldsymbol{\theta}_{l-1}$ are two snapshots from the same training process, and so the similarity between them is higher than that in multi-generation T-S optimization. This makes the second term on the right-hand side of Eqn~\ref{Eqn:SnapshotDistillation} degenerate and, consequently, its contribution to the gradient that $\boldsymbol{\theta}_{l-1}$ receives for updating itself is considerably changed.

\renewcommand{\colwidthA}{1.04cm}
\newcommand{\colwidthB}{2.00cm}
\begin{table}[!btp]
\centering{
\setlength{\tabcolsep}{0.16cm}
\begin{tabular}{|L{\colwidthA}|R{\colwidthB}|R{\colwidthB}|R{\colwidthB}|}
\hline
{}                   &
$\mathbb{M}_{\#0}^\mathrm{T}$ &
$\mathbb{M}_{\#75}^\mathrm{T}$ &
$\mathbb{M}_{\#150}^\mathrm{T}$ \\
\hline\hline
$\mathbb{M}_{\#0}$   & $78.13$ &                       $-$ &                       $-$ \\
\hline
$\mathbb{M}_{\#75}$  & $78.18$  & $78.02$ &                       $-$ \\
\hline
$\mathbb{M}_{\#150}$ & $77.67$  & $77.58$  & $77.47$  \\
\hline
\end{tabular}}
\caption{Classification error rates ($\%$) on CIFAR100 with different T-S similarities. All these models are trained for $300$ epochs, and all numbers are the average of two individual runs. The first row (self) shows the accuracies of standard models (no T-S optimization), and in the following rows, when $\mathbb{M}_{\#E_1}^\mathrm{T}$ teaches $\mathbb{M}_{\#E_2}$, they share the first $\min\!\left\{E_1,E_2\right\}$ common epochs. Some T-S pairs that are probabilistically identical, so only one of them is tested (see Section~\ref{Approach:Principles:Difference} for details).}
\label{Tab:Similarity}
\vspace{-0.0cm}
\end{table}

We evaluate the impact of T-S similarity using the $100$-layer DenseNet~\cite{huang2017densely} on the CIFAR100 dataset~\cite{krizhevsky2009learning}. All models are trained with the cosine annealing learning rate policy~\cite{loshchilov2016sgdr} for a total of $300$ epochs. Detailed settings are elaborated in Section~\ref{Experiments:CIFAR100}. To construct T-S pairs with different similarities, we first perform a complete training process containing $300$ standard epochs and starting from scratch, and denote the final model by $\mathbb{M}_{\#300}$. Then, we take the snapshots at $150$, $75$ and $0$ (scratch) epochs, and denote them by $\mathbb{M}_{\#150}$, $\mathbb{M}_{\#75}$ and $\mathbb{M}_{\#0}$, respectively, with the number after $\#$ indicating the number of elapsed epochs. Then, we continue training these snapshots with the same configurations (mini-batch size, learning rates, {\em etc.}) but different randomization which affects the sampled mini-batch in each iteration and the data augmentation performed at each training sample. These models are denoted by $\mathbb{M}_{\#150}^\mathrm{T}$, $\mathbb{M}_{\#75}^\mathrm{T}$ and $\mathbb{M}_{\#0}^\mathrm{T}$, respectively, where the superscript $\mathrm{T}$ implies being used as a teacher model, and each number after $\#$ indicates the number of common epochs shared with $\mathbb{M}_{\#300}$. All these teacher models have exactly $300$ epochs.

Now, we use these models to teach the intermediate snapshots, {\em i.e.}, $\mathbb{M}_{\#150}$, $\mathbb{M}_{\#75}$ and $\mathbb{M}_{\#0}$. When $\mathbb{M}_{\#E_1}^\mathrm{T}$ is used to teach $\mathbb{M}_{\#E_2}$, their common part, {\em i.e.}, the first ${E_0}={\min\!\left\{E_1,E_2\right\}}$ epochs are preserved, {\em i.e.}, the first $E_0$ epochs used Eqn~\eqref{Eqn:StandardLoss} and the remaining $300-E_0$ epochs used Eqn~\eqref{Eqn:StudentLoss}. Results are summarized in Table~\ref{Tab:Similarity}. Note that from a probabilistic perspective, $\mathbb{M}_{\#150}^\mathrm{T}$, $\mathbb{M}_{\#75}^\mathrm{T}$ and $\mathbb{M}_{\#0}^\mathrm{T}$ are identical to each other in classification accuracy, and from the previous part we expect them to provide the same teaching ability. We start with observing their behavior when $\mathbb{M}_{\#0}$ is the student. This case degenerates to a two-generation T-S optimization. Since all teachers are probabilistically identical, we only evaluate one of these pairs, reporting a $78.13\%$ accuracy which is higher than the baseline (the average of $\mathbb{M}_{\#150}^\mathrm{T}$, $\mathbb{M}_{\#75}^\mathrm{T}$ and $\mathbb{M}_{\#0}^\mathrm{T}$ is $77.65\%$). However, when $\mathbb{M}_{\#75}$ is the student, $\mathbb{M}_{\#0}^\mathrm{T}$ serves as a better teacher because it does not share the first $75$ epochs with $\mathbb{M}_{\#75}$. This offers a larger difference between teacher and student and, consequently, produces better classification performance ($78.18\%$ vs. $78.02\%$). When $\mathbb{M}_{\#150}$ is chosen to be the student, this phenomenon preserves, {\em i.e.}, T-S optimization prefers a larger difference between teacher and student.

\iffalse
A side discovery of this experiments comes from the comparison between the T-S pairs in which $\mathbb{M}_{\#0}^\mathrm{T}$ is the teacher, and thus three students, $\mathbb{M}_{\#0}$, $\mathbb{M}_{\#75}$ and $\mathbb{M}_{\#150}$, have the same similarity with the teacher but different starting points of T-S optimization. We find that $\mathbb{M}_{\#0}$ and $\mathbb{M}_{\#75}$ converge to similar performance, but $\mathbb{M}_{\#150}$ is significantly worse. From these results, we learn the lesson that T-S optimization have a higher impact in the iterations with smaller learning rates.
\fi

\subsubsection{Principle \#3: Secondary Information}
\label{Approach:Principles:SecondaryInformation}

The last factor, also being the one that was most studied before, is how knowledge is delivered from teacher to student. There are two arguments, both of which suggesting that a smoother teacher signal preserves richer information, but they differ from each other in the way of achieving this goal. The {\em distillation} algorithm~\cite{hinton2015distilling} used a temperature term $T$ to smooth both input and output scores, and the {\em tolerant teacher} algorithm~\cite{yang2018knowledge} trained a less confident teacher by adding a regularization term in the first generation ({\em a.k.a.} the patriarch), and this strategy was verified the advantageous over the non-regularized version~\cite{furlanello2018born}.

In the context of snapshot distillation, we follow~\cite{hinton2015distilling} to divide the teacher signal (in {\em logits}, the neural responses before the softmax layer) by a temperature coefficient ${T}>{1}$. In the framework of knowledge distillation, the student signals should also be softened before the KL divergence is computed with the teacher signals. The reason is that, the student with a shallow architecture is not capable of perfectly fitting the outputs of the teacher with a deep architecture, and thus matching the soft versions of their outputs is a more rational choice. The aim of knowledge distillation is to match the outputs, forcing the student to predict what the teacher predicts as much as possible. However, our goal is to generate secondary information in T-S optimization, instead of matching. As a result, we do not divide the student signal by $T$. This strategy also aligns with Eqn~\ref{Eqn:StandardLoss} used in the very first iterations ({\em i.e.}, no teacher signals are provided). In experiments, we observe a faster convergence as well as consistent accuracy gain -- see Section~\ref{Experiments:CIFAR100} for detailed numbers. We name it as {\em asymmetric distillation}.

\iffalse
It remains to adjust the coefficient of mixing two signals, {\em i.e.}, $\lambda_l$ and $1-\lambda_l$ in Eqn~\eqref{Eqn:SnapshotDistillation}, in order to cooperate with our new distillation strategy.
\fi

\subsubsection{Summary}
\label{Approach:Principles:Summary}

Summarizing the above three principles, we present our solution to improve the performance of SD. We partition the entire training process with $L$ iterations into $K$ mini-generations with $L_1,L_2\ldots,L_K$ iterations, respectively, and ${{\sum_{k=1}^K}L_k}={L}$. The last iteration in each mini-generation serves as the teacher of all iterations in the next mini-generation. This is to say, there are $K-1$ teachers. The first teacher is the snapshot at ${L_1'}={L_1}$ iterations, the second one at ${L_2'}={L_1+L_2}$ iterations, and the last one at ${{L_{K-1}'}}={L_1+L_2\ldots+L_{K-1}}$ iterations. We have: \begin{equation}
\label{Eqn:TeacherIteration}
{c_l}={\max\left\{L_k',L_k'<l\right\}}.
\end{equation}
For ${l}\leqslant{L_1'}$, we define ${c_l}={0}$ for later convenience, and in this case ${\lambda_{l}^{S}}=1$, $\lambda_{l}^{T}=0$ and Eqn~\eqref{Eqn:SnapshotDistillation} degenerates to Eqn~\eqref{Eqn:StandardLoss}. Following {\bf Principle \#2}, we shall assume that the iterations right after each teacher have large learning rates, in order to ensure the sufficient difference between the teacher and student models. Meanwhile, according to {\bf Principle \#1}, the teacher itself should be good, which implies that the iterations before each teacher have small learning rates, making the network converge to an acceptable state. To satisfy both conditions, we require the learning rates within each mini-generation to start from a large value and gradually go down. In practice, we use the cosine annealing strategy~\cite{loshchilov2016sgdr} which was verified to converge better:
\begin{equation}
\label{Eqn:LearningRate}
{\gamma_l}={\frac{1}{2}\alpha_{k_l}\times\left[1+\cos\!\left(\frac{l-L_{k_l-1}'}{L_{k_l}'-L_{k_l-1}'}\cdot\pi\right)\right]}.
\end{equation}
Here, $k_l$ is the index of mini-generation of $l$, and $\alpha_{k_l}$ is the starting learning rate at the beginning of this mini-generation (often set to be large). Finally, we follow Section~\ref{Approach:Principles:SecondaryInformation} to use asymmetric distillation in order to satisfy {\bf Principle \#3}.

\subsection{Discussions}
\label{Approach:Discussions}

If we set ${L_1}={L_2}=\ldots={L_K}$ and switch off asymmetric distillation, the above solution degenerates to snapshot ensemble (SE)~\cite{huang2018snapshot}. In experiments, we compare these two approaches under the same setting, and find that both approaches work well on CIFAR100 (SD reports better results), but on ILSVRC2012, SD achieves higher accuracy over the baseline while SE does not\footnote{The SE paper~\cite{huang2018snapshot} reported a higher accuracy on ResNet50, but it was compared to the baseline with the stepwise learning rate policy, not the cosine annealing policy that should be the direct baseline. The latter baseline is more than $1\%$ higher than the former, and also outperforms SE.}. This is arguably because CIFAR100 is relatively simple, so that the original setting ($L$ iterations) are over-sufficient for convergence, and thus reducing the number of iterations of each mini-generation does not cause significant accuracy drop. ILSVRC2012, however, is much more challenging and thus convergence becomes a major drawback of both SD and SE. SD, with the extra benefit brought by T-S optimization, bridges this gap and outperforms the baseline.

Also, note that the above solution is only one choice. Under the generalized framework (Algorithm~\ref{Alg:SnapshotDistillation}) and following these three principles, other training strategies can be explored, {\em e.g.}, using super-convergence~\cite{smith2017super} to alleviate the drawback of weaker convergence. These options will be studied in the future.

\section{Experiments}
\label{Experiments}

\subsection{The CIFAR100 Dataset}
\label{Experiments:CIFAR100}

\renewcommand{\colwidthA}{0.8cm}
\renewcommand{\colwidthB}{1.4cm}
\begin{table*}[!htp]
\centering{
\setlength{\tabcolsep}{0.08cm}
\begin{tabular}{|l||C{\colwidthA}|C{\colwidthA}||C{\colwidthB}|C{\colwidthB}|C{\colwidthB}|C{\colwidthB}|C{\colwidthB}|C{\colwidthB}||C{1.4cm}|C{1.0cm}|}
\hline
Backbone & Alg. & $T$ & $\mathbb{M}_{\#L_1}$ & $\mathbb{M}_{\#L_2}$ & $\mathbb{M}_{\#L_3}$ & $\mathbb{M}_{\#L_4}$ & {\em best} & {\em ensemble} & \multicolumn{2}{c|}{SOTA} \\
\hline\hline
\multirow{4}{*}{ResNet20} & {\bf BL} & N/A & $-$ & $-$ & $-$ & $33.57$ & $33.57$ & $-$ & \multirow{2}{*}{\em Year} & \multirow{2}{*}{--|}\\
\cline{2-9}
{} & {\bf SE} & N/A & $36.17$ & $33.36$ & $32.98$ & $32.66$ & $32.54$ & $30.86$ & {} & {}\\
\cline{2-11}
{} & {\bf SD} & $2$ & $36.17$ & $33.78$ & $32.98$ & $32.31$ & $32.31$ & $32.08$ & \multirow{2}{*}{$2016$~\cite{zagoruyko2016wide}} & \multirow{2}{*}{$19.25$}\\
\cline{2-9}
{} & {\bf SD} & $3$ & $36.17$ & $33.69$ & $32.24$ & $\mathbf{31.97}$ & $\mathbf{31.76}$ & $\mathbf{30.76}$ & {} & {}\\
\hline\hline
\multirow{4}{*}{ResNet32} & {\bf BL} & N/A & -- & -- & -- & $31.61$ & $31.61$ & -- & \multirow{2}{*}{$2017$~\cite{zhang2017interleaved}} & \multirow{2}{*}{$19.25$}\\
\cline{2-9}
{} & {\bf SE} & N/A & $33.78$ & $32.15$ & $31.41$ & $30.74$ & $30.51$ & $28.93$ & {} & {}\\
\cline{2-11}
{} & {\bf SD} & $2$ & $33.78$ & $32.07$ & $31.05$ & $30.67$ & $30.57$ & $29.80$ & \multirow{2}{*}{$2017$~\cite{zhong2017random}} & \multirow{2}{*}{$17.73$}\\
\cline{2-9}
{} & {\bf SD} & $3$ & $33.78$ & $31.52$ & $30.64$ & $\mathbf{30.32}$ & $\mathbf{30.16}$ & $\mathbf{28.71}$ & {} & {}\\
\hline\hline
\multirow{4}{*}{ResNet56} & {\bf BL} & N/A & -- & -- & -- & $30.23$ & $29.94$ & -- & \multirow{2}{*}{$2017$~\cite{xie2017aggregated}} & \multirow{2}{*}{$17.31$}\\
\cline{2-9}
{} & {\bf SE} & N/A & $32.85$ & $31.60$ & $30.45$ & $29.68$ & $29.55$ & $27.93$ & {} & {}\\
\cline{2-11}
{} & {\bf SD} & $2$ & $32.85$ & $30.47$ & $29.72$ & $\mathbf{29.29}$ & $\mathbf{29.22}$ & $28.11$ & \multirow{2}{*}{$2017$~\cite{huang2017densely}} & \multirow{2}{*}{$17.18$}\\
\cline{2-9}
{} & {\bf SD} & $3$ & $32.85$ & $30.82$ & $29.55$ & $29.37$ & $29.28$ & $\mathbf{27.74}$ & {} & {}\\
\hline\hline
\multirow{4}{*}{ResNet110} & {\bf BL} & N/A & -- & -- & -- & $28.77$ & $28.53$ & -- & \multirow{2}{*}{$2017$~\cite{han2017deep}} & \multirow{2}{*}{$17.01$}\\
\cline{2-9}
{} & {\bf SE} & N/A & $31.89$ & $29.81$ & $29.07$ & $28.27$ & $28.09$ & $26.45$ & {} & {}\\
\cline{2-11}
{} & {\bf SD} & $2$ & $31.89$ & $29.84$ & $28.71$ & $\mathbf{27.71}$ & $\mathbf{27.52}$ & $27.19$ & \multirow{2}{*}{$2017$~\cite{zhang2017mixup}} & \multirow{2}{*}{$16.80$}\\
\cline{2-9}
{} & {\bf SD} & $3$ & $31.89$ & $29.22$ & $28.37$ & $27.87$ & $27.75$ & $\mathbf{26.19}$ & {} & {}\\
\hline\hline
\multirow{4}{*}{DenseNet100} & {\bf BL} & N/A & -- & -- & -- & $22.49$ & $22.00$ & -- & \multirow{2}{*}{$2017$~\cite{dong2017eraserelu}} & \multirow{2}{*}{$16.53$}\\
\cline{2-9}
{} & {\bf SE} & N/A & $24.31$ & $22.76$ & $22.16$ & $22.18$ & $22.00$ & $\mathbf{19.63}$ & {} & {}\\
\cline{2-11}
{} & {\bf SD} & $2$ & $24.31$ & $23.10$ & $22.06$ & $21.78$ & $21.59$ & $20.27$ & \multirow{2}{*}{$2017$~\cite{gastaldi2017shake}} & \multirow{2}{*}{$15.85$}\\
\cline{2-9}
{} & {\bf SD} & $3$ & $24.31$ & $23.19$ & $21.60$ & $\mathbf{21.17}$ & $\mathbf{21.17}$ & $19.71$ & {} & {}\\
\hline\hline
\multirow{4}{*}{DenseNet190} & {\bf BL} & N/A & -- & -- & -- & $16.82$ & $16.69$ & -- & \multirow{2}{*}{$2018$~\cite{furlanello2018born}} & \multirow{2}{*}{$14.90^{\ast}$}\\
\cline{2-9}
{} & {\bf SE} & N/A & $18.98$ & $18.12$ & $16.95$ & $\mathbf{16.84}$ & $16.70$ & $\mathbf{15.70}$ & {} & {}\\
\cline{2-11}
{} & {\bf SD} & $2$ & $18.98$ & $17.48$ & $16.32$ & $18.02$ & \textcolor{red}{$\mathbf{16.06}$} & $15.72$ & \multirow{2}{*}{$2018$~\cite{yang2018knowledge}} & \multirow{2}{*}{$14.47^{\ast}$}\\
\cline{2-9}
{} & {\bf SD} & $3$ & $18.98$ & $17.67$ & $16.95$ & $18.65$ & $16.33$ & $15.92$ & {} & {}\\
\hline
\end{tabular}}
\caption{
CIFAR100 classification errors ($\%$) obtained by different network backbones. Regarding the algorithm option, {\bf BL} indicate the {\em baseline} model trained with cosine annealing learning rates, {\bf SE} indicates {\em snapshot ensemble} with the same learning rate policy as {\bf SD} during the entire training process. $T$ is the temperature term. We report the accuracy at the end of each mini-generation, at the {\em best} epoch, and obtained from model ensemble ($\mathbb{M}_{\#L_1}$ through $\mathbb{M}_{\#L_4}$), respectively. The logits of $\mathbb{M}_{\#L_{k}}$ are multiplied by $T^{k-1}$ for ensemble of \textbf{SD}. Among the state-of-the-art (SOTA) methods, an asterisk indicates that model ensemble was used to achieve the corresponding error rate. In addition, \cite{gastaldi2017shake} used complicated data augmentation to achieve an error rate of $15.85\%$ -- we just applied standard data augmentation.}
\vspace{-0.0cm}
\label{Tab:CIFAR100}
\end{table*}

\subsubsection{Settings and Baselines}
\label{Experiments:CIFAR100:Settings}

We first evaluate SD on the CIFAR100 dataset~\cite{krizhevsky2009learning}, a low-resolution ($32\times32$) dataset containing $60\rm{,}000$ RGB images. These images are split into a training set of $50\rm{,}000$ images and a testing set of $10\rm{,}000$ images, and in both of them, images are uniformly distributed over all $100$ classes ($20$ superclasses each of which contains $5$ fine-level classes). We do not perform experiments on the CIFAR10 dataset because it does not contain fine-level visual concepts, and thus the benefit brought by T-S optimization is not significant (this was also observed in~\cite{furlanello2018born} and analyzed in~\cite{yang2018knowledge}).

We investigate two groups of baseline models. The first group contains standard deep ResNets~\cite{he2016deep} with $20$, $32$, $56$ and $110$ layers. Let the total number of layers be $A$, then ${A'}={\left(A-2\right)/6}$ is the number of residual blocks in each stage. Given a $32\times32$ input image, a $3\times3$ convolution is first performed without changing its spatial resolution. Three stages followed, each of which has $A'$ residual blocks (two $3\times3$ convolutions summed up with an identity connection). Batch normalization~\cite{ioffe2015batch} and ReLU activation~\cite{nair2010rectified} are applied after each convolutional layer. The spatial resolution changes in the three stages ($32\times32$, $16\times16$ and $8\times8$), as well as the number of channels ($16$, $32$ and $64$). An average pooling layer is inserted after each of the first two stages. The network ends with global average-pooling followed by a fully-connected layer with $100$ outputs. The second group has two DenseNets~\cite{huang2017densely} with $100$ and $190$ layers, respectively. These networks share the similar architecture with the ResNets, but the building blocks in each stage are densely-connected, with the output of each block concatenated to the accumulated feature vector and fed into the next block. The base feature length and growth rate are $24$ and $12$ for DenseNet100, and $80$ and $40$ for DenseNet190.

Following the conventions, we train all these networks from scratch. We use the standard Stochastic Gradient Descent (SGD) with a weight decay of $0.0001$ and a Nesterov momentum of $0.9$. In ResNets, we train the network for $164$ epochs with a mini-batch size of $128$ and a base learning rate of $0.1$. In DenseNets, we train the network for $300$ epochs with a mini-batch size of $64$ and a base learning rate of $0.1$. The cosine annealing learning rate~\cite{loshchilov2016sgdr} is used, in order to make fair comparison between the baseline and SD. In the training process, standard data-augmentation is used, {\em i.e.}, each image is symmetrically-padded with a $4$-pixel margin on each of the four sides. In the enlarged $40\times40$ image, a subregion with $32\times32$ pixels is randomly cropped and flipped with a probability of $0.5$. We do not use any data augmentation in the testing stage. 

To apply SD, we evenly partition the entire training process into $4$ mini-generations, {\em i.e.}, ${K}={4}$. For ResNets, we have ${L_1}={41}$, ${L_2}={82}$ and ${L_3}={123}$, and for DenseNets, ${L_1}={75}$, ${L_2}={150}$ and ${L_3}={225}$. The same learning rate ${\alpha_k}={0.1}$ is used at the beginning of each mini-generation, and decayed following Eqn~\eqref{Eqn:LearningRate}. We use an asymmetric distillation strategy (Section ~\ref{Approach:Principles:SecondaryInformation}) with ${T}={2}$ and ${T}={3}$, respectively. In Eqn~\ref{Eqn:SnapshotDistillation}, we set ${\lambda_l^\mathrm{S}}={1+1/T}$ and ${\lambda_l^\mathrm{T}}={1}$ to approximately balance two sources of gradients in their magnitudes~\cite{hinton2015distilling}.
\vspace{0cm}

\subsubsection{Quantitative Results and Analysis}
\label{Experiments:CIFAR100:Results}

Results are summarized in Table~\ref{Tab:CIFAR100}. Towards fair comparison, for different instances of the same backbone, network weights are initialized in the same way, although randomness during the training process ({\em e.g.}, data shuffle and augmentation) is not unified. In addition, the first mini-generation ($\mathbb{M}_{\#L_1}$, no T-S optimization) is shared between SE (snapshot ensemble) and SD.

We first observe that SD brings consistent accuracy gain for all models, regardless of network backbones, and surpassing both the baseline and SE. In DenseNet190, the most powerful baseline, SD with ${T}={2}$ achieves an error rate of $16.06\%$ at the best epoch, which is competitive among the state-of-the-arts (all of which reported the best epoch). Moreover, in terms of model ensemble from $\mathbb{M}_{\#L_1}$ through $\mathbb{M}_{\#L_4}$, SD provides comparable numbers to SE, although we emphasize that SD focuses on optimizing a single model while SE, with weaker single models, requires ensemble to improve classification accuracy. Another explanation comes from the optimization policy of SD. By introducing a teacher signal to optimize each student, different snapshots in SD tend to share a higher similarity than SE, and this is the reason that SD reports a smaller accuracy gain from a single model to model ensemble.

Another important topic to discuss is how asymmetric distillation impacts T-S optimization, for which we show several evidences. With a temperature term ${T}>{1}$, the student tends to become smoother, {\em i.e.}, the entropy of the class distribution is larger. However, as shown in~\cite{furlanello2018born} and~\cite{yang2018knowledge}, T-S optimization achieves satisfying performance via finding a balancing point between certainty and uncertainty, so, as the latter gradually increases, we can observe a {\em peak} in classification accuracy. In DenseNet190 with ${T}={2}$, this peak appears during the third mini-generation which achieves the lowest error rate at $16.06\%$, but the final error rate goes up $18.02\%$. A similar phenomenon also appears in DenseNet100 with ${T}={4}$, which also achieves the lowest error at the third mini-generation (the lowest error of $21.26\%$ vs. the last error $21.86\%$), and in ResNets with ${T}>{5}$. This reveals that the optimal temperature term is closely related to the network backbone. For a deeper backbone ({\em e.g.}, DenseNet190) which itself has a strong ability of fitting data, we use a smaller $T$ to introduce less soft labels, decreasing the ambiguity.

%we can use a smaller $T$ so that the supervision signal from the teacher is weakened.

\subsection{The ILSVRC2012 Dataset}
\label{Experiments:ILSVRC2012}

\subsubsection{Settings and Baselines}
\label{Experiments:ILSVRC2012:Settings}

We now investigate a much more challenging dataset, ILSVRC2012~\cite{russakovsky2015imagenet}, which is a popular subset of the ImageNet database~\cite{deng2009imagenet}. It contains $1.3\mathrm{M}$ training images and $50\mathrm{K}$ testing images, all of which are high-resolution, covering $1\rm{,}000$ object classes in total. The distribution over classes is approximately uniform in the training set and and strictly uniform in the testing set.

We use deep ResNets~\cite{he2016deep} with $101$ and $152$ layers. They share the same overall design with the ResNets used for CIFAR100, but in each residual block, there is a so-called bottleneck structure which, in order to accelerate, compresses the number of channels by $3/4$ and later recovers the original number. Each input image has a size of $224\times224$. After the first $7\times7$ convolutional layer with a stride of $2$ and a $3\times3$ max-pooling layer, four main stages follow with different numbers of blocks (ResNet101: $3,4,23,3$; ResNet152: $3,8,36,3$). The spatial resolutions in these four stages are $56\times56$, $28\times28$, $14\times14$ and $7\times7$, and the number of channels are $256$, $512$, $1\rm{,}024$ and $2\rm{,}048$, respectively. Three max-pooling layers are inserted between these four stages. The network ends with global average-pooling followed by a fully-connected layer with $1\rm{,}000$ outputs.

We follow the conventions to configure the training parameters. The standard Stochastic Gradient Descent (SGD) with a weight decay of $0.0001$ and a Nesterov momentum of $0.9$ is used. In a total of $90$ epochs, the mini-batch size is fixed to be $256$. We still use the cosine annealing learning rate~\cite{loshchilov2016sgdr} starting with $0.1$. A series of data-augmentation techniques~\cite{szegedy2015going} are applied in training to alleviate over-fitting, including rescaling and cropping the image, randomly mirroring and rotating (slightly) the image, changing its aspect ratio and performing pixel jittering. In the testing stage, the standard single-center-crop is used.

To apply SD, we set ${K}={2}$ which partitions the training process into two equal sections (each has $45$ epochs). The reason of using a smaller $K$ (compared to CIFAR experiments) is that on ILSVRC2012 with high-resolution images and more complex semantics, it is much more difficult to guarantee convergence with a fewer number of iterations within each mini-generation. Regarding the temperature term, we fix ${T}={2}$. Other settings are the same as in the CIFAR experiments.

\subsubsection{Quantitative Results}
\label{Experiments:ILSVRC2012:Results}

\renewcommand{\colwidthA}{0.8cm}
\renewcommand{\colwidthB}{1.2cm}
\begin{table}[!htp]
\centering{
\setlength{\tabcolsep}{0.08cm}
\begin{tabular}{|l|C{\colwidthA}|C{\colwidthB}|C{\colwidthB}|C{\colwidthB}|C{\colwidthB}|}
\hline
\multirow{2}{*}{Backbone} & \multirow{2}{*}{Alg.} & \multicolumn{2}{c|}{$\mathbb{M}_{\#L_1}$} & \multicolumn{2}{c|}{$\mathbb{M}_{\#L_2}$} \\
\cline{3-6}
{} & {} & Top-$1$ & Top-$5$ & Top-$1$ & Top-$5$ \\
\hline\hline
ResNet101 & {\bf BL} & $-$ & $-$ & $21.62$ & $5.80$ \\
\hline
ResNet101 & {\bf SE} & $22.94$ & $6.51$ & $22.14$ & $6.07$ \\
\hline
ResNet101 & {\bf SD} & $22.94$ & $6.51$ & $\mathbf{21.25}$ & $\mathbf{5.55}$ \\
\hline\hline
ResNet152 & {\bf BL} & $-$ & $-$ & $21.17$ & $5.66$ \\
\hline
ResNet152 & {\bf SE} & $22.56$ & $6.44$ & $21.84$ & $5.84$ \\
\hline
ResNet152 & {\bf SD} & $22.56$ & $6.44$ & $\mathbf{20.93}$ & $\mathbf{5.55}$ \\
\hline
\end{tabular}}
\caption{ILSVRC2012 classification errors ($\%$) obtained by different network backbones. Regarding the algorithm option, {\bf BL} indicate the {\em baseline} model trained with cosine annealing learning rates, and {\bf SD} snapshot distillation with ${T}={2}$.}
\label{Tab:ILSVRC2012}
\vspace{-0.2cm}
\end{table}

Experimental results are summarized in Table~\ref{Tab:ILSVRC2012}. To make fair comparison, the first mini-generation of {\bf SE} and {\bf SD} are shared, and thus the only difference lies in whether the teacher signal is provided in the second mini-generation. We can see that the performance of {\bf SE} is consistently worse than that of both {\bf BL} and {\bf SD}. Even if the two mini-generations of {\bf SE} are fused, the error rates ($21.66\%$ on ResNet101 and $21.19\%$ on ResNet152) are slightly higher than {\bf BL}. This reveals that reducing the number of training epochs harms the ability of learning from a challenging dataset such as ILSVRC2012. Our approach, {\bf SD} applies a teacher signal as a remedy, so that the training process becomes more efficient especially under a limited number of iterations.

In addition, SD achieves consistent accuracy gain over the baseline in terms of both top-$1$ and top-$5$ error rates. On ResNet101, the top-$1$ and top-$5$ errors drop by $0.37\%$ and $0.25\%$ absolutely, or $1.71\%$ and $4.31\%$ relatively; on ResNet152, the top-$1$ and top-$5$ errors drop by $0.26\%$ and $0.11\%$ absolutely, or $1.23\%$ and $1.94\%$ relatively. These improvement seems small, but we emphasize that (i) to the best of our knowledge, this is the first time that a model achieves higher accuracy on ILSVRC2012 with T-S optimization within one generation; and (ii) these accuracy gain transfers well to other visual recognition tasks, as shown in the next subsection.

%\footnote{The original paper of snapshot ensemble~\cite{huang2018snapshot} reported slightly higher accuracy on ResNet18, but we note that it was compared to the baseline obtained using the step-wise learning rate policy. In a fair comparison to the actual baseline with the cosine annealing policy, it was outperformed by more than }

%The settings follow those in the main article (see Section~4.2.1). Here we add snapshot ensemble ({\bf SE})~\cite{huang2018snapshot} into comparison.

%Consequently, {\bf SD} produces stronger models beyond the baseline.

We plot the curves of both the baseline and SD in the training process of ResNet152. We can see that, in the second mini-generation, SD achieves a higher training error but a lower testing error, {\em i.e.}, the gap between training and testing accuracies becomes smaller, which aligns with our motivation that T-S optimization alleviates over-fitting.

\newcommand{\figurewidthA}{5.2cm}
\newcommand{\figurewidthB}{2.93cm}
\begin{figure}[!t]
\centering
\includegraphics[height=3.8cm]{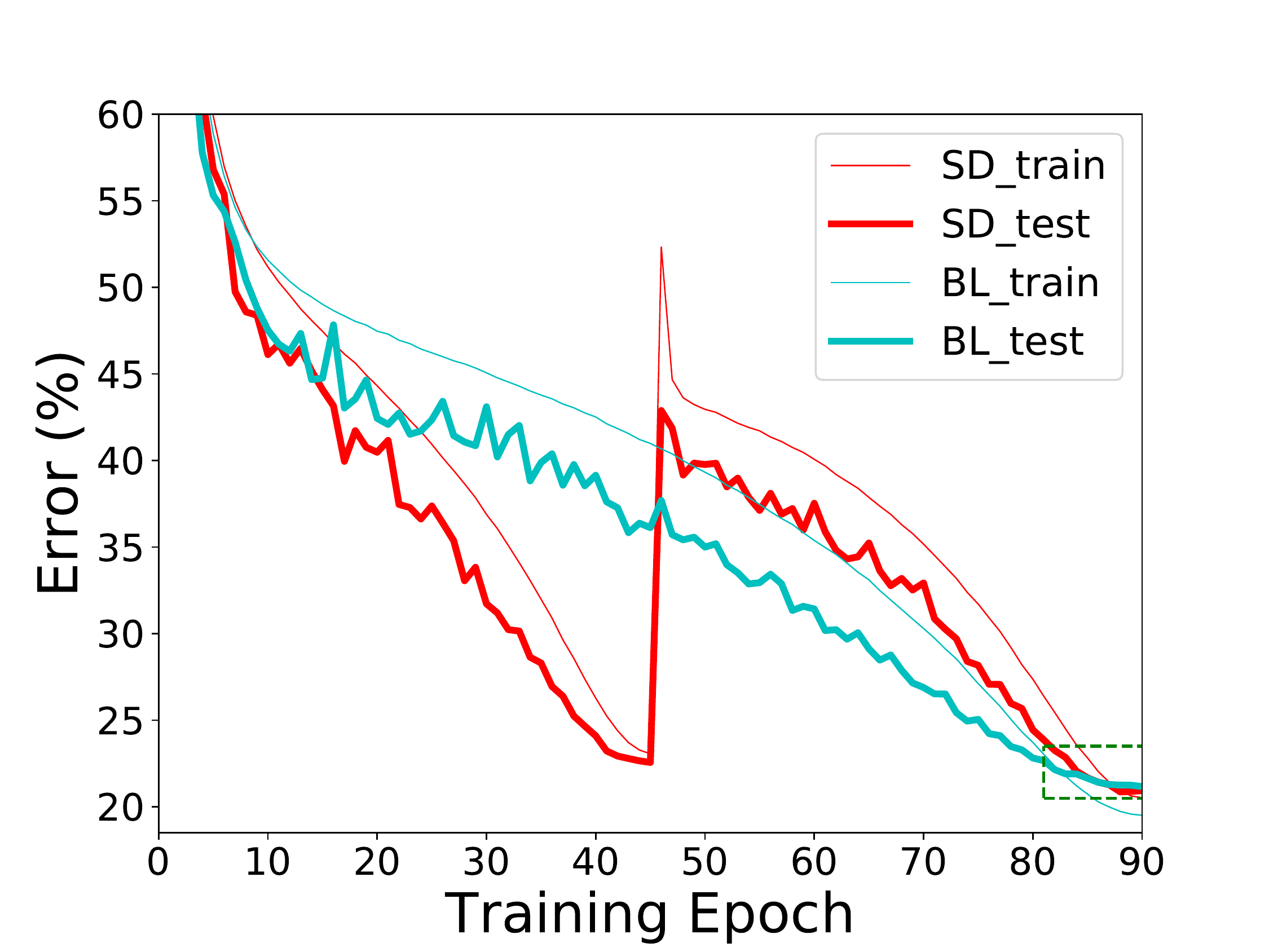}
\includegraphics[height=3.8cm]{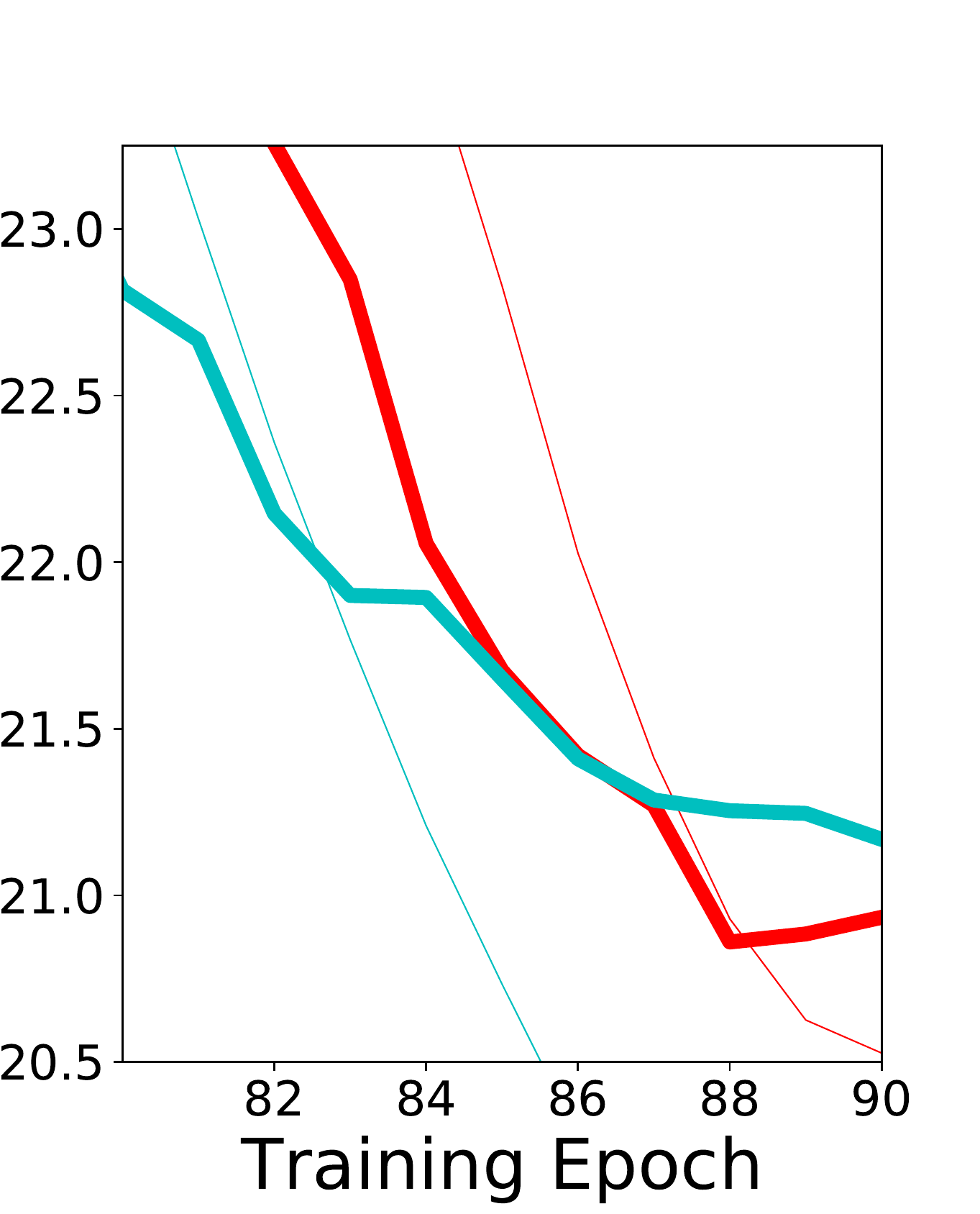}
\caption{The training and testing curves of ResNet152.}
\label{Fig:Curves}
\end{figure}

\subsection{Transfer Experiments}
\vspace{-0.2cm}
\label{Experiments:Transfer}

Last but not least, we fine-tune the models pre-trained on ILSVRC2012 to the object detection and semantic segmentation tasks in the PascalVOC dataset~\cite{everingham2010pascal}, a widely used benchmark in computer vision. The most powerful models, {\em i.e.}, the baseline and SD versions of ResNet152, are transferred using a standard approach, which preserves the network backbone (all layers before the final pooling layer), and introduces a network head known as Faster R-CNN~\cite{ren2015faster} for object detetion, and DeepLab-v3~\cite{chen2017rethinking} for semantic segmentation.

\renewcommand{\colwidthA}{2.2cm}
\renewcommand{\colwidthB}{2.2cm}
\begin{table}[!t]
\centering
\begin{tabular}{|L{\colwidthA}|C{\colwidthB}|C{\colwidthB}|}
\hline
Backbone           & mAP @ 2007       & mIOU @ 2012      \\
\hline\hline
ResNet152-{\bf BL} & $73.49$          & $77.53$          \\
\hline
ResNet152-{\bf SD} & $\mathbf{74.93}$ & $\mathbf{77.97}$ \\
\hline
\end{tabular}
\caption{PascalVOC object detection (2007, mAP, $\%$) and semantic segmentation (2012, mIOU, $\%$) results, both obtained by fine-tuning the pre-trained deep networks on ILSVRC2012 with Faster R-CNN~\cite{ren2015faster} and DeepLab-v3~\cite{chen2017rethinking}.}
\label{Tab:Detection}
\vspace{-0.2cm}
\end{table}

This model is fine-tuned in an end-to-end manner. For object detection on PascalVOC 2007, $5\rm{,}011$ training images are fed into the network through $10$ epochs with a mini-batch size of $16$. We start a learning rate of $0.01$ and divide it by $10$ after $8$ epochs. For semantic segmentation on PascalVOC 2012, $10\rm{,}582$ training images~\cite{hariharan2011semantic} are fed into the network through $50$ epochs with a mini-batch size of $8$. We use ``poly'' learning rate policy where the initial learning rate is $0.007$ and the power is $0.9$. Results in terms of mAP and mIOU are summarized in Table~\ref{Tab:Detection}. One can see that, the model with a higher accuracy on ILSVRC2012 also works better in both tasks, {\em i.e.}, the benefit brought by SD preserves after fine-tuning. Also, we emphasize that SD, providing the same network architecture but being stronger, does not require any additional costs in transfer learning, which claims its potential applications in a wide range of vision problems.

\section{Conclusions}
\label{Conclusions}

In this paper, we present a framework named snapshot distillation (SD), which finishes teacher-student (T-S) optimization within one generation. To the best of our knowledge, this goal was never achieved before. The key contribution is to take teacher signals from the previous iterations of the same training process, and discuss on three principles that impact the performance of SD. The final solution is easy to implement yet efficient to carry out. With around $1/3$ extra training time, SD boosts the classification accuracy of several baseline models on CIFAR100 and ILSVRC2012 consistently, and the performance gain persists after the trained model is fine-tuned on other vision tasks, {\em e.g.}, object detection, semantic segmentation.

Our research reduces the basic unit of T-S optimization from a complete generation to a mini-generation which is composed of a number of iterations. The essential difficulty that prevents us from further partitioning this unit is the requirement of T-S difference. We believe there exists, though not yet found, a way of eliminating this constraint so that the basic unit can be even smaller, {\em e.g.}, one single iteration. If this is achieved, we may directly integrate supervision from the previous iteration into the current iteration, obtaining a new loss function in which the teacher signal appears as a term of higher-order gradients. We leave this topic for future research.

\vspace{0.2cm}
\noindent{\bf Acknowledgements}\quad This paper is supported by NSF award CCF-1317376 and ONR award N00014-15-1-2356. We thank Siyuan Qiao, Huiyu Wang and Chenxi Liu who provided insight and expertise to improve the research.
{\small
\bibliographystyle{ieee}
\bibliography{egbib}

\begin{thebibliography}{10}\itemsep=-1pt

\bibitem{akata2016label}
Z.~Akata, F.~Perronnin, Z.~Harchaoui, and C.~Schmid.
\newblock Label-embedding for image classification.
\newblock {\em IEEE Transactions on Pattern Analysis and Machine Intelligence},
  38(7):1425--1438, 2016.

\bibitem{bagherinezhad2018label}
H.~Bagherinezhad, M.~Horton, M.~Rastegari, and A.~Farhadi.
\newblock Label refinery: Improving imagenet classification through label
  progression.
\newblock {\em arXiv preprint arXiv:1805.02641}, 2018.

\bibitem{chen2016deeplab}
L.~C. Chen, G.~Papandreou, I.~Kokkinos, K.~Murphy, and A.~L. Yuille.
\newblock Deeplab: Semantic image segmentation with deep convolutional nets,
  atrous convolution, and fully connected crfs.
\newblock In {\em International Conference on Learning Representations}, 2016.

\bibitem{chen2017rethinking}
L.~C. Chen, G.~Papandreou, F.~Schroff, and H.~Adam.
\newblock Rethinking atrous convolution for semantic image segmentation.
\newblock {\em arXiv preprint arXiv:1706.05587}, 2017.

\bibitem{chen2016net2net}
T.~Chen, I.~Goodfellow, and J.~Shlens.
\newblock Net2net: Accelerating learning via knowledge transfer.
\newblock In {\em International Conference on Learning Representations}, 2016.

\bibitem{deng2010what}
J.~Deng, A.~C. Berg, K.~Li, and L.~Fei-Fei.
\newblock What does classifying more than 10,000 image categories tell us?
\newblock In {\em European Conference on Computer Vision}, 2010.

\bibitem{deng2009imagenet}
J.~Deng, W.~Dong, R.~Socher, L.~Li, K.~Li, and L.~Fei-Fei.
\newblock Imagenet: A large-scale hierarchical image database.
\newblock In {\em Computer Vision and Pattern Recognition}, 2009.

\bibitem{donahue2014decaf}
J.~Donahue, Y.~Jia, O.~Vinyals, J.~Hoffman, N.~Zhang, E.~Tzeng, and T.~Darrell.
\newblock Decaf: A deep convolutional activation feature for generic visual
  recognition.
\newblock In {\em International Conference on Machine Learning}, 2014.

\bibitem{dong2017eraserelu}
X.~Dong, G.~K., K.~Zhan, and Y.~Yang.
\newblock Eraserelu: a simple way to ease the training of deep convolution
  neural networks.
\newblock {\em arXiv preprint arXiv:1709.07634}, 2017.

\bibitem{everingham2010pascal}
M.~Everingham, L.~Van~Gool, C.~K.~I. Williams, J.~Winn, and A.~Zisserman.
\newblock The pascal visual object classes (voc) challenge.
\newblock {\em International Journal of Computer Vision}, 88(2):303--338, 2010.

\bibitem{furlanello2018born}
T.~Furlanello, Z.~C. Lipton, L.~Itti, and A.~Anandkumar.
\newblock Born again neural networks.
\newblock In {\em International Conference on Machine Learning}, 2018.

\bibitem{gastaldi2017shake}
X.~Gastaldi.
\newblock Shake-shake regularization.
\newblock {\em arXiv preprint arXiv:1705.07485}, 2017.

\bibitem{girshick2015fast}
R.~Girshick.
\newblock Fast r-cnn.
\newblock In {\em Computer Vision and Pattern Recognition}, 2015.

\bibitem{girshick2014rich}
R.~Girshick, J.~Donahue, T.~Darrell, and J.~Malik.
\newblock Rich feature hierarchies for accurate object detection and semantic
  segmentation.
\newblock In {\em Computer Vision and Pattern Recognition}, 2014.

\bibitem{guo2017calibration}
C.~Guo, G.~Pleiss, Y.~Sun, and K.~Q. Weinberger.
\newblock On calibration of modern neural networks.
\newblock In {\em International Conference on Machine Learning}, 2017.

\bibitem{han2017deep}
D.~Han, J.~Kim, and J.~Kim.
\newblock Deep pyramidal residual networks.
\newblock In {\em Computer Vision and Pattern Recognition}, 2017.

\bibitem{hariharan2011semantic}
B.~Hariharan, P.~Arbel{\'a}ez, L.~Bourdev, S.~Maji, and J.~Malik.
\newblock Semantic contours from inverse detectors.
\newblock In {\em International Conference on Computer Vision}, 2011.

\bibitem{he2016deep}
K.~He, X.~Zhang, S.~Ren, and J.~Sun.
\newblock Deep residual learning for image recognition.
\newblock In {\em Computer Vision and Pattern Recognition}, 2016.

\bibitem{hinton2015distilling}
G.~Hinton, O.~Vinyals, and J.~Dean.
\newblock Distilling the knowledge in a neural network.
\newblock {\em arXiv preprint arXiv:1503.02531}, 2015.

\bibitem{hu2018squeeze}
J.~Hu, L.~Shen, and G.~Sun.
\newblock Squeeze-and-excitation networks.
\newblock In {\em Computer Vision and Pattern Recognition}, 2018.

\bibitem{huang2018snapshot}
G.~Huang, Y.~Li, G.~Pleiss, Z.~Liu, J.~E. Hopcroft, and K.~Q. Weinberger.
\newblock Snapshot ensembles: Train 1, get m for free.
\newblock In {\em International Conference on Learning Representations}, 2018.

\bibitem{huang2017densely}
G.~Huang, Z.~Liu, K.~Q. Weinberger, and L.~van~der Maaten.
\newblock Densely connected convolutional networks.
\newblock In {\em Computer Vision and Pattern Recognition}, 2017.

\bibitem{ioffe2015batch}
S.~Ioffe and C.~Szegedy.
\newblock Batch normalization: Accelerating deep network training by reducing
  internal covariate shift.
\newblock In {\em International Conference on Machine Learning}, 2015.

\bibitem{krizhevsky2009learning}
A.~Krizhevsky and G.~Hinton.
\newblock Learning multiple layers of features from tiny images.
\newblock 2009.

\bibitem{krizhevsky2012imagenet}
A.~Krizhevsky, I.~Sutskever, and G.~E. Hinton.
\newblock Imagenet classification with deep convolutional neural networks.
\newblock In {\em Advances in Neural Information Processing Systems}, 2012.

\bibitem{lecun2015deep}
Y.~LeCun, Y.~Bengio, and G.~E. Hinton.
\newblock Deep learning.
\newblock {\em Nature}, 521(7553):436, 2015.

\bibitem{liu2018progressive}
C.~Liu, B.~Zoph, J.~Shlens, W.~Hua, L.~J. Li, L.~Fei-Fei, A.~L. Yuille,
  J.~Huang, and K.~Murphy.
\newblock Progressive neural architecture search.
\newblock In {\em European Conference on Computer Vision}, 2018.

\bibitem{long2015fully}
J.~Long, E.~Shelhamer, and T.~Darrell.
\newblock Fully convolutional networks for semantic segmentation.
\newblock In {\em Computer Vision and Pattern Recognition}, 2015.

\bibitem{loshchilov2016sgdr}
I.~Loshchilov and F.~Hutter.
\newblock Sgdr: Stochastic gradient descent with warm restarts.
\newblock {\em arXiv preprint arXiv:1608.03983}, 2016.

\bibitem{nair2010rectified}
V.~Nair and G.~E. Hinton.
\newblock Rectified linear units improve restricted boltzmann machines.
\newblock In {\em International Conference on Machine Learning}, 2010.

\bibitem{pereyra2017regularizing}
G.~Pereyra, G.~Tucker, J.~Chorowski, {\L}.~Kaiser, and G.~Hinton.
\newblock Regularizing neural networks by penalizing confident output
  distributions.
\newblock {\em arXiv preprint arXiv:1701.06548}, 2017.

\bibitem{perronnin2010improving}
F.~Perronnin, J.~Sanchez, and T.~Mensink.
\newblock Improving the fisher kernel for large-scale image classification.
\newblock In {\em European conference on computer vision}, 2010.

\bibitem{razavian2014cnn}
A.~S. Razavian, H.~Azizpour, J.~Sullivan, and S.~Carlsson.
\newblock Cnn features off-the-shelf: an astounding baseline for recognition.
\newblock In {\em Computer Vision and Pattern Recognition}, 2014.

\bibitem{ren2015faster}
S.~Ren, K.~He, R.~Girshick, and J.~Sun.
\newblock Faster r-cnn: Towards real-time object detection with region proposal
  networks.
\newblock In {\em Advances in Neural Information Processing Systems}, 2015.

\bibitem{romero2015fitnets}
A.~Romero, N.~Ballas, S.~E. Kahou, A.~Chassang, C.~Gatta, and Y.~Bengio.
\newblock Fitnets: Hints for thin deep nets.
\newblock In {\em International Conference on Learning Representations}, 2014.

\bibitem{russakovsky2015imagenet}
O.~Russakovsky, J.~Deng, H.~Su, J.~Krause, S.~Satheesh, S.~Ma, Z.~Huang,
  A.~Karpathy, A.~Khosla, M.~Bernstein, et~al.
\newblock Imagenet large scale visual recognition challenge.
\newblock {\em International Journal of Computer Vision}, 115(3):211--252,
  2015.

\bibitem{simonyan2015very}
K.~Simonyan and A.~Zisserman.
\newblock Very deep convolutional networks for large-scale image recognition.
\newblock In {\em International Conference on Learning Representations}, 2015.

\bibitem{smith2017super}
L.~N. Smith and N.~Topin.
\newblock Super-convergence: Very fast training of residual networks using
  large learning rates.
\newblock {\em arXiv preprint arXiv:1708.07120}, 2017.

\bibitem{srivastava2014dropout}
N.~Srivastava, G.~E. Hinton, A.~Krizhevsky, I.~Sutskever, and R.~Salakhutdinov.
\newblock Dropout: A simple way to prevent neural networks from overfitting.
\newblock {\em Journal of Machine Learning Research}, 15(1):1929--1958, 2014.

\bibitem{szegedy2015going}
C.~Szegedy, W.~Liu, Y.~Jia, P.~Sermanet, S.~Reed, D.~Anguelov, D.~Erhan,
  V.~Vanhoucke, A.~Rabinovich, et~al.
\newblock Going deeper with convolutions.
\newblock In {\em Computer Vision and Pattern Recognition}, 2015.

\bibitem{szegedy2016rethinking}
C.~Szegedy, V.~Vanhoucke, S.~Ioffe, J.~Shlens, and Z.~Wojna.
\newblock Rethinking the inception architecture for computer vision.
\newblock In {\em Computer Vision and Pattern Recognition}, 2016.

\bibitem{tarvainen2017mean}
A.~Tarvainen and H.~Valpola.
\newblock Mean teachers are better role models: Weight-averaged consistency
  targets improve semi-supervised deep learning results.
\newblock In {\em Advances in Neural Information Processing Systems}, 2017.

\bibitem{verma2012learning}
N.~Verma, D.~Mahajan, S.~Sellamanickam, and V.~Nair.
\newblock Learning hierarchical similarity metrics.
\newblock In {\em Computer Vision and Pattern Recognition}, 2012.

\bibitem{wang2014learning}
J.~Wang, T.~Leung, C.~Rosenberg, J.~Wang, J.~Philbin, B.~Chen, Y.~Wu, et~al.
\newblock Learning fine-grained image similarity with deep ranking.
\newblock In {\em Computer Vision and Pattern Recognition}, 2014.

\bibitem{wu2017hierarchical}
C.~Wu, M.~Tygert, and Y.~LeCun.
\newblock Hierarchical loss for classification.
\newblock {\em arXiv preprint arXiv:1709.01062}, 2017.

\bibitem{xie2017genetic}
L.~Xie and A.~Yuille.
\newblock Genetic cnn.
\newblock In {\em International Conference on Computer Vision}, 2017.

\bibitem{xie2017aggregated}
S.~Xie, R.~Girshick, P.~Doll{\'a}r, Z.~Tu, and K.~He.
\newblock Aggregated residual transformations for deep neural networks.
\newblock In {\em Computer Vision and Pattern Recognition}, 2017.

\bibitem{xie2015holistically}
S.~Xie and Z.~Tu.
\newblock Holistically-nested edge detection.
\newblock In {\em International Conference on Computer Vision}, 2015.

\bibitem{yang2018knowledge}
C.~Yang, L.~Xie, S.~Qiao, and A.~L. Yuille.
\newblock Knowledge distillation in generations: More tolerant teachers educate
  better students.
\newblock {\em arXiv preprint arXiv:1805.05551}, 2018.

\bibitem{yim2017gift}
J.~Yim, D.~Joo, J.~Bae, and J.~Kim.
\newblock A gift from knowledge distillation: Fast optimization, network
  minimization and transfer learning.
\newblock In {\em Computer Vision and Pattern Recognition}, 2017.

\bibitem{zagoruyko2016wide}
S.~Zagoruyko and N.~Komodakis.
\newblock Wide residual networks.
\newblock {\em arXiv preprint arXiv:1605.07146}, 2016.

\bibitem{zhang2018image}
C.~Zhang, J.~Cheng, and Q.~Tian.
\newblock Image-level classification by hierarchical structure learning with
  visual and semantic similarities.
\newblock {\em Information Sciences}, 422:271--281, 2018.

\bibitem{zhang2017mixup}
H.~Zhang, M.~Cisse, Y.~N. Dauphin, and D.~Lopez-Paz.
\newblock mixup: Beyond empirical risk minimization.
\newblock {\em arXiv preprint arXiv:1710.09412}, 2017.

\bibitem{zhang2017interleaved}
T.~Zhang, G.~J. Qi, B.~Xiao, and J.~Wang.
\newblock Interleaved group convolutions.
\newblock In {\em Computer Vision and Pattern Recognition}, 2017.

\bibitem{zhang2017deep}
Y.~Zhang, T.~Xiang, T.~M. Hospedales, and H.~Lu.
\newblock Deep mutual learning.
\newblock {\em arXiv preprint arXiv:1706.00384}, 2017.

\bibitem{zhong2017random}
Z.~Zhong, L.~Zheng, G.~Kang, S.~Li, and Y.~Yang.
\newblock Random erasing data augmentation.
\newblock {\em arXiv preprint arXiv:1708.04896}, 2017.

\bibitem{zoph2017neural}
B.~Zoph and Q.~V. Le.
\newblock Neural architecture search with reinforcement learning.
\newblock In {\em International Conference on Learning Representations}, 2017.

\end{thebibliography}
}

\end{document}